\documentclass{article}

\usepackage[final]{corl_2024} %
\usepackage{graphicx}
\usepackage{booktabs}
\usepackage{hyperref}

\title{STEVE-Audio: Expanding the Goal Conditioning Modalities of Embodied Agents in Minecraft}

\author{
  Nicholas Lenzen \\
  Bielefeld University, Germany \\
  \texttt{nicholas.lenzen@uni-bielefeld.de} \\
  \And
  Amogh Raut \\
  Indian Institute of Technology BHU, Varanasi, India \\
  \texttt{amoghprashant.raut.cd.mat19@itbhu.ac.in} \\
  \And
  Andrew Melnik \\
  Bremen University, Germany \\ 
}

\begin{document}
\maketitle

\begin{abstract}
Recently, the STEVE-1 approach has been introduced as a method for training generative agents to follow instructions in the form of latent CLIP embeddings.
In this work, we present a methodology to extend the control modalities
by learning a mapping from new input modalities to the latent goal space of the agent. We apply our approach to the challenging Minecraft domain, and extend the goal conditioning 
to include the audio modality. The resulting audio-conditioned agent is able to perform on a comparable level to the original text-conditioned and visual-conditioned agents. Specifically, we create an Audio-Video CLIP foundation model for Minecraft and an audio prior network which together map audio samples to the latent goal space of the STEVE-1 policy. Additionally, we highlight the tradeoffs that occur when conditioning on different modalities. Our training code, evaluation code, and Audio-Video CLIP foundation model for Minecraft are made open-source to help foster further research into multi-modal generalist sequential decision-making agents.For additional resources, including code and demonstrations, please visit our project website: \url{https://sites.google.com/itbhu.ac.in/steve-audio}.

\end{abstract}


\section{Introduction}

Recent works have shown that we can train
generalist sequential decision-making agents
\cite{chen2021decision, brohan2023rt, baker2022video, lifshitz2023steve1}.
In particular, Lifshitz et al. \cite{lifshitz2023steve1} introduced the STEVE-1 approach for creating generative instruction-following agents to follow short-horizon instructions without being trained on a specific set of tasks, by learning to follow instructions represented as latent vectors in a CLIP \cite{radford2021learning} embedding space. However, this approach produces generative agents that are limited to following instructions in the specific input modalities of the CLIP model.

Creating agents which are increasingly multi-modal enables the creation of singular agents that can leverage the advantages provided by each prompting modality. 
Modalities like audio and video often co-occur together,
which facilitates the collection of internet-scale datasets with relatively low effort \cite{baker2022video,lifshitz2023steve1}. Thus, the ability to create increasingly multi-modal agents or extending the prompting modalities of existing agents is an important area for investigation, as multi-modal agents can leverage the advantages provided by each prompting modality.

In this work, we introduce a methodology for extending the prompting modalities of generative agents trained to follow instructions.
We apply our methodology to the Minecraft domain by enhancing the STEVE-1 Minecraft agent \cite{lifshitz2023steve1} to follow audio prompts in addition to its original text and visual prompting modalities. Our findings show that the audio-conditioned agent surpasses both the text-conditioned and visual-conditioned agents in certain tasks, although it encounters difficulties in others. This suggests that tradeoffs can arise when transitioning between modalities. For instance, our results indicate that audio prompting may demand less prompt engineering; however, there are also tasks that cannot be fully conveyed through audio alone.

Generative agents created using the STEVE-1 approach \cite{lifshitz2023steve1} learn to follow instructions represented as goal embeddings in the latent space of a CLIP model \cite{radford2021learning}. Our method extends the prompting modalities of such agents without retraining the policy by 1) training a new CLIP model where one of the modalities is the new prompting modality, 2) learning a prior which maps from the new CLIP latent space to the latent space of the CLIP model originally used to train the STEVE-1-based generative agent. Thus, given a sample instruction in the new modality, we generate an embedding for the sample using the new CLIP model, then map this embedding to the latent goal space using the prior, and condition the STEVE-1 policy on the generated latent goal vector to generate behavior using keyboard/mouse controls in Minecraft.

Our main contributions are as follows:

\begin{itemize}
    \item We introduce a method for extending the prompting modalities of generative agents.
    \item We apply our methodology to the Minecraft domain by extending the STEVE-1 Minecraft agent to follow audio prompts.
    \item We discuss and demonstrate the tradeoffs that occur when switching between different prompting modalities in the Minecraft domain.
    \item Our training code, evaluation code, and Audio-Video CLIP foundation model for Minecraft are made open-source to help foster further research into multi-modal generalist sequential decision-making agents.
    \item We release a 600-hour Audio-Video dataset of Minecraft gameplay sourced from the internet and augmented with task-specific gameplay.
\end{itemize}

\begin{figure}[t]
     \centering
     \includegraphics[width=1.0\linewidth]{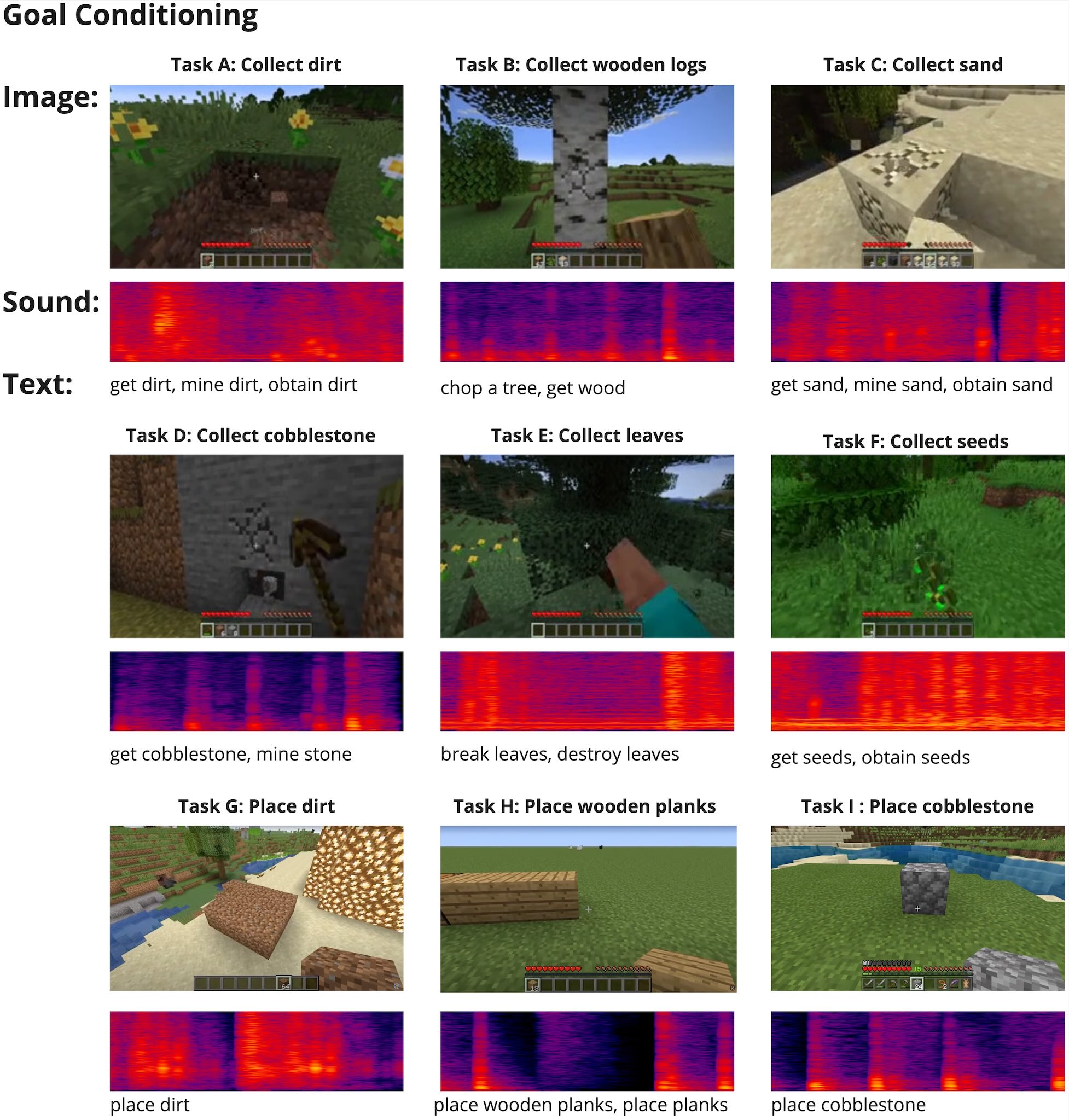}
     \caption{
Examples of evaluation tasks within the Minecraft environment, showcasing the observed ego-centric views as the agent works to achieve the corresponding objectives.
}
     \label{tasks}
\end{figure}

\section{Related work}

\paragraph{Minecraft for AI} Minecraft has become a popular environment for testing AI agents due to its open-ended nature, providing a broad spectrum of tasks (e.g., \cite{baker2022video,fan2022minedojo,guss2019minerl,lifshitz2023steve1,malato2022behavioral,malato2024zero,melnik2021critic,milani2023towards,wang2023voyager,wang2023describe,wang2023jarvis}). Frameworks like MineRL \cite{guss2019minerl,milani2023towards} and MineDojo \cite{fan2022minedojo} facilitated this trend by providing the necessary tools to run AI agents in Minecraft. Notable models include MineCLIP \cite{fan2022minedojo}, which integrates Minecraft video and text prompts into a shared latent space, VPT \cite{baker2022video}, a generative model for behavior in Minecraft, STEVE-1 \cite{lifshitz2023steve1}, which generates behavior based on visual and text goals. These models are specifically designed to operate within the Minecraft environment, enabling AI to perform a variety of complex tasks. These range from simple activities like collecting blocks and defeating enemies to more complex challenges such as crafting items, which require accomplishing numerous secondary tasks. 

\paragraph{Multi-modal Decision-Making} Several prior works explored the use of LLMs in creating Minecraft agents that can follow instructions \cite{cai2023groot, wang2023jarvis, wang2023describe, zhou2024minedreamer}. These works typically use LLMs to make high-level plans that are then executed by lower-level RL \cite{wang2023describe} or scripted \cite{prismarine2022mineflayer} policies. JARVIS-1 \cite{wang2023jarvis} is an open-world agent that uses a memory-augmented multimodal language model to achieve planning and control in Minecraft, capable of completing over 200 tasks. Groot \cite{cai2023groot} is an agent that learns to follow open-ended instructions by watching gameplay videos, achieving advanced goal specification and control. MineDreamer \cite{zhou2024minedreamer} is an embodied agent that uses a Chain-of-Imagination mechanism to effectively follow diverse, abstract, and sequential instructions in Minecraft. MP5 \cite{qin2023mp5} is a multimodal embodied Minecraft agent that decomposes complex tasks, designs context-aware plans, and performs goal-conditioned actions in process-dependent and context-dependent tasks. 
MUTEX \cite{shah2023mutex} introduces a transformer-based approach for policy learning from multimodal task specifications, enabling agents to follow instructions and goals across six modalities, including video, images, text, and speech. AVLEN \cite{paul2022avlen}introduces a multi-modal hierarchical reinforcement learning agent that uses audio-visual cues and natural language assistance to localize audio events and navigate 3D environment.

\paragraph{CLIP for Combining Modalities} The introduction of the CLIP model \cite{radford2021learning} led to the creation of a multitude of CLIP-based models that combine different modalities into a common latent space e.g., AudioCLIP \cite{guzhov2022audioclip} is an extension of the CLIP model that integrates audio processing capabilities with text and image modalities, achieving results in Environmental Sound Classification. CLIP4VLA \cite{ruan2023accommodating} accommodates the audio modality, enabling Vision-Language-Audio multimodal processing for video retrieval and captioning tasks. CLASP \cite{rana2023contrastive} extends the CLIP by incorporating behaviour-text alignment. MineCLIP \cite{fan2022minedojo} is a contrastive video-language model pre-trained on massive YouTube database. Correlating different modalities made it possible to create models like DALL-E 2 \cite{ramesh2022hierarchical} or the aforementioned STEVE-1, where the generation of one modality is conditioned on another modality for a given input.

\paragraph{CLIP for Reinforcement Learning} Another possible application of this CLIP method is to use it for reward shaping in reinforcement learning \cite{bach2020learn}. This can be done by measuring how similar the embedding of the current episode is to an embedding created from a goal that the agent should achieve. This approach is used in the domain of Minecraft \cite{fan2022minedojo} as well as other domains including robotics \cite{dang2023clip,baumli2023vision,sontakke2024roboclip, rana2023contrastive}.
Off-the-shelf vision-language models (VLMs), such as the CLIP family, can be used as effective sources of rewards for training reinforcement learning agents to achieve a variety of language goals in rich, open-ended environments \cite{baumli2023vision}.
CLIP-Motion \cite{dang2023clip} is a method for learning reward functions for robotic actions by leveraging a CLIP-based model to process consecutive observations and generalize across various robotic tasks. RoboCLIP \cite{sontakke2024roboclip} is an online imitation learning method that uses a single video or textual demonstration to generate rewards without manual reward function design, significantly improving zero-shot performance in robot manipulation tasks by leveraging pretrained Video-and-Language Models (VLMs).

\section{Method}

In STEVE-1 \cite{lifshitz2023steve1}, Lifshitz et al. introduce an approach to train generative instruction-following agents by learning to follow instructions represented as goal embeddings in the latent space of a CLIP model \cite{radford2021learning}. This approach consists of two parts, the policy and a prior. The policy extends the VPT model \cite{baker2022video} in order to enable it to fullfill goals in the form of visual MineCLIP \cite{fan2022minedojo} embeddings. In order to enable the agent to follow instructions given through text, the text encoder of MineCLIP is employed to encode the text prompt, which is then mapped onto a visual MineCLIP embedding by the prior network which is a CVAE \cite{kingma2022autoencoding,NIPS2015_8d55a249} trained to sample latent visual MineCLIP embeddings conditioned on a text embedding. However, the resulting agent can only follow the two modalities of the employed CLIP model (in the case of STEVE-1, the two modalities of MineCLIP are text and video). 

To address this problem, we introduce a methodology for extending the prompting modality of generative agents trained with the approach proposed by Lifshitz et al. \cite{lifshitz2023steve1}. We train a new CLIP model where one of the modalities is the new prompting modality. Then, we learn a mapping from the new CLIP latent space to the CLIP latent space originally used to train the generative agent. Thus, to follow instructions in the new prompting modality, we encode the instruction with the corresponding encoder for the new CLIP model, pass it through the prior to obtain a latent goal, and then pass this latent goal to the policy of the agent trained using the approach from \cite{lifshitz2023steve1}. 

We apply our methodology to the challenging Minecraft domain, where we extend the STEVE-1 agent to follow audio prompts by creating an Audio-Visual CLIP model for Minecraft. We pick video as the other modality for our CLIP model since both audio and video are naturally occurring modalities, which means that we can scalably train the model using unlabelled Minecraft videos.

\subsection{Audio-Video CLIP Foundation Model for Minecraft}

\begin{figure}[t]
     \centering
     \includegraphics[width=\linewidth]{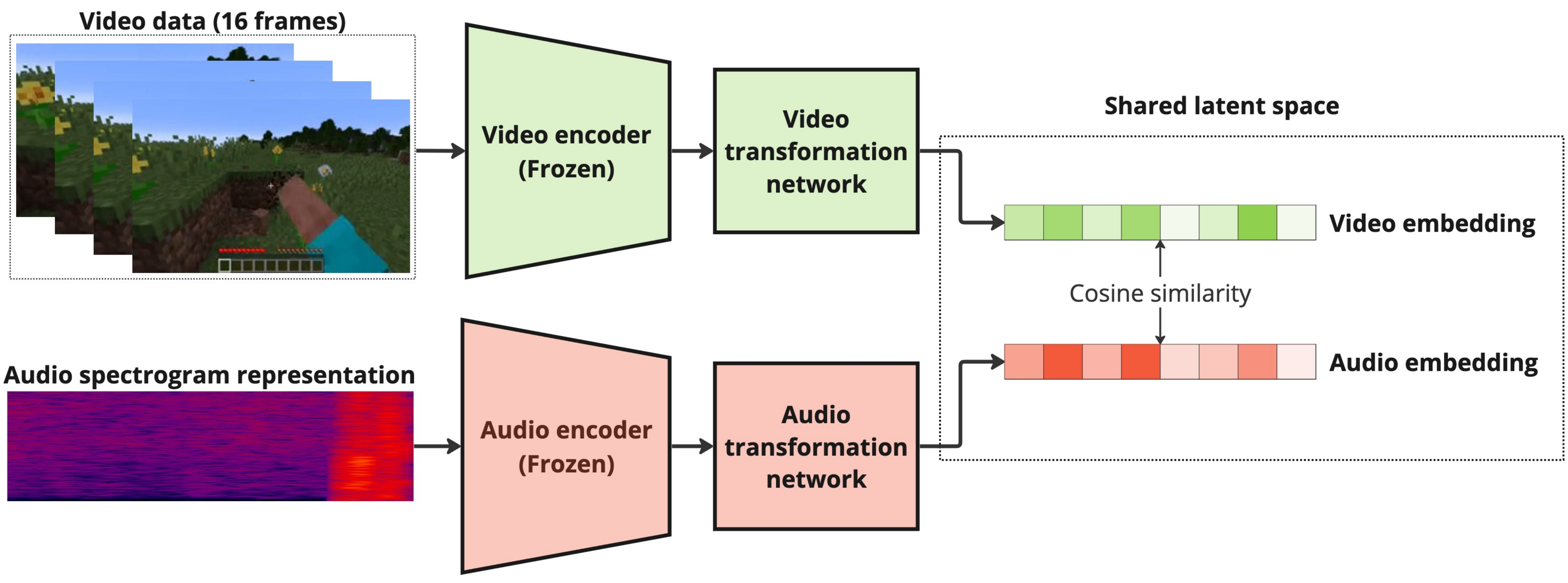}
     \caption{
Our architecture for the Audio-Video CLIP model learns a shared latent space by jointly training the audio and video transformation networks, which are versions of the mapping network used by StyleGAN 3 \cite{Karras2021}. We utilized frozen pretrained MineCLIP \cite{fan2022minedojo} model for the video encoder and frozen pretrained Audio Spectrogram Transformer for the audio encoder \cite{gong21b_interspeech}.
         }
     \label{clip_model}
\end{figure}

Our proposed Audio-Video CLIP foundation model consists of a frozen video-encoder and frozen audio-encoder on top of which we train non-linear transformation networks to transform the encoder embeddings into a new, shared latent space. We train this model on the Audio-Video dataset as described in the next subsection. Below, we outline the details of the video encoder, audio encoder, and transformation networks. See Figure \ref{clip_model} for a visualization of the Audio-Video CLIP model architecture. 

\paragraph{Video Encoder} We use the pretrained video encoder from the MineCLIP model \cite{fan2022minedojo} which maps 16-frames of video input and text to a joint embedding space. Specifically, the MineCLIP video encoder consists of a frame-wise image encoder, which creates an embedding for each of the 16 input frames followed by a temporal pooling network. This network pools the 16 frame-wise embeddings into a single embedding for the whole video input. 
We use the attention-based pooling version of MineCLIP. See \cite{fan2022minedojo} for more details about MineCLIP.

\paragraph{Audio Encoder} We use the pretrained Audio Spectrogram Transfomer (AST) \cite{gong21b_interspeech} model which was originally trained to classify audio spectrograms into different categories (i.e., speech, vehicle, musical instrument, etc.). We embed Minecraft audio samples using AST by first computing the corresponding spectrogram representation of the audio sample, then passing this spectrogram to the AST encoder, and using the logits of the model as audio embeddings. %

\paragraph{Transformation Networks} The MineCLIP video encoder and AST audio encoder weights are frozen, but we also train non-linear transformation networks on top of these models to transform the latent embeddings from each encoder to a new shared latent space. These transformation networks serve two purposes. First, training one transformation network is necessary to ensure that the latent vectors have the same size. Second, training both transformation networks allows our Audio-Video CLIP model to learn its own latent space. For example, if we only added a learned transformation network to the audio encoder, then the objective of the audio encoder's transformation network would be to mimic the latent space of the MineCLIP video encoder, which could potentially restrict the expressivity of the learned embeddings. We found that the best-performing architecture for the transformation network is an upscaled version of the mapping network used by StyleGAN 3 \cite{Karras2021, melnik2024face}. Specifically, we increase the layer count from eight to ten and the hidden dimension from $512$ to $1024$. Further, we use cosine similarity as the similarity metric between both the two latent vectors, as in the original CLIP paper \cite{radford2021learning}. Both transformation networks map their input dimensionality (527 for audio embeddings and 512 for video embeddings) to a 512-vector. Training goal of the Audio-Video CLIP model is to maximize the cosine similarity of matching audio-video pairs, while simultaneously minimizing the cosine similarity of not matching pairs found in a training batch. This is achieved by employing a contrastive learning scheme as was used to train the original CLIP model \cite{radford2021learning}.

\subsection{Audio-Video Dataset} 
\label{dataset}

To train the Audio-Video CLIP foundation model, we collect a dataset of Minecraft video and accompanying audio (where the audio is not overlayed with music or commentary). Most of the dataset is sourced from YouTube videos of Minecraft gameplay without commentary, where we cut off the first two minutes of gameplay since many videos start with a short, irrelevant introduction sequence. We used 25 hours of such data to train the Audio-Video CLIP model which we use to generate our results. However, the full dataset which we plan to release contains about 600 hours of data. See appendix for more details about the dataset.

\subsection{Training the Audio-Video CLIP Foundation Model}

Our training scheme is the same as the original CLIP model \cite{radford2021learning}, except that we do not use a cosine scheduler \cite{loshchilovstochastic} to decay the learning rate. As previously mentioned, we freeze the weights of both encoder models during training and only update the weights of the transformation networks, which we train for 100 epochs. 
See Table \ref{table:dataset-metrics} and \ref{table:train-hyperparams} in appendix \ref{appendix} for a list of hyperparameters.

\subsection{Extending STEVE-1 to Condition on Audio Prompts}
\label{audio-steve}

\begin{figure}
    \centering
    \includegraphics[width=\linewidth]{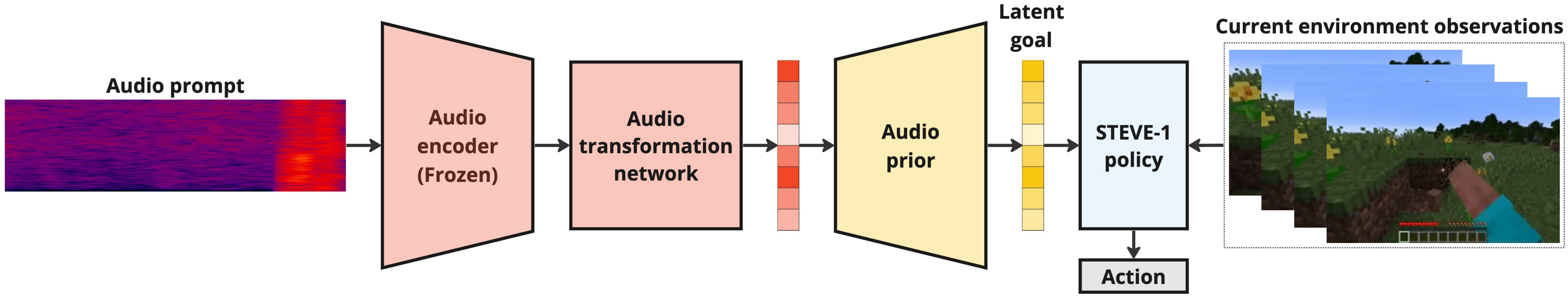}
    \caption{
    Our architecture for audio prompting of the STEVE-1 agent \cite{lifshitz2023steve1}.
    }
    \label{steve_pipeline}
\end{figure}

We aim to extend the STEVE-1 agent to condition on audio prompts. To that end, we train the Audio-Video CLIP foundation model which learns a new shared latent space for audio and video modalities. However, we cannot directly condition the STEVE-1 policy on embeddings in this new latent space, as it is different to the MineCLIP latent space that the policy was trained on. Thus, to condition the STEVE-1 policy on audio, we must train a prior which maps audio embeddings from the Audio-Video CLIP model to visual MineCLIP embeddings. Note that this prior does not map audio embeddings to visual embeddings from our Audio-Video CLIP model, but rather to visual embedding from the MineCLIP model (which the STEVE-1 policy was trained to follow).

The architecture of our prior, which maps audio embeddings to visual MineCLIP embeddings, is similar to the prior in STEVE-1 \cite{lifshitz2023steve1}, which maps text MineCLIP embeddings to visual MineCLIP embeddings. It is implemented as a CVAE \cite{kingma2022autoencoding,NIPS2015_8d55a249} where the encoder and decoder are both two-layer MLPs with a hidden dimension of $256$ and layer normalization between layers. Thus, to condition STEVE-1 on an audio sample, we first compute the audio embedding using the audio encoder from our Audio-Video CLIP model. Then, we use the prior to map this audio embedding into the latent goal space of the policy (which is the visual MineCLIP embedding space), which we can use to condition the STEVE-1 policy and generate instruction-following behavior with keyboard/mouse controls in Minecraft. See Figure \ref{steve_pipeline} for the architecture of STEVE-Audio.

\subsection{Evaluation} \label{evaluation}

Following the programmatic evaluation methodology from Baker et al. \cite{baker2022video} and Lifshitz et al. \cite{lifshitz2023steve1}, we evaluate the performance of the audio-conditioned STEVE-1 agent \cite{lifshitz2023steve1} on a set of short-horizon item-collection tasks in Minecraft (collecting dirt, wooden logs, seeds, sand, cobblestone, and leaves). We compare the performance of the audio, text, and visual conditioned versions of STEVE-1 with minimal prompt engineering. Each task was evaluated on 10 different seeds for 2 minutes each (2400 timesteps at 20 frames-per-second). See Figure \ref{tasks} for the goal conditioning tasks with corresponding audio and text prompts.

\begin{figure}[t]
     \centering
    
     \includegraphics[width=0.8\linewidth]{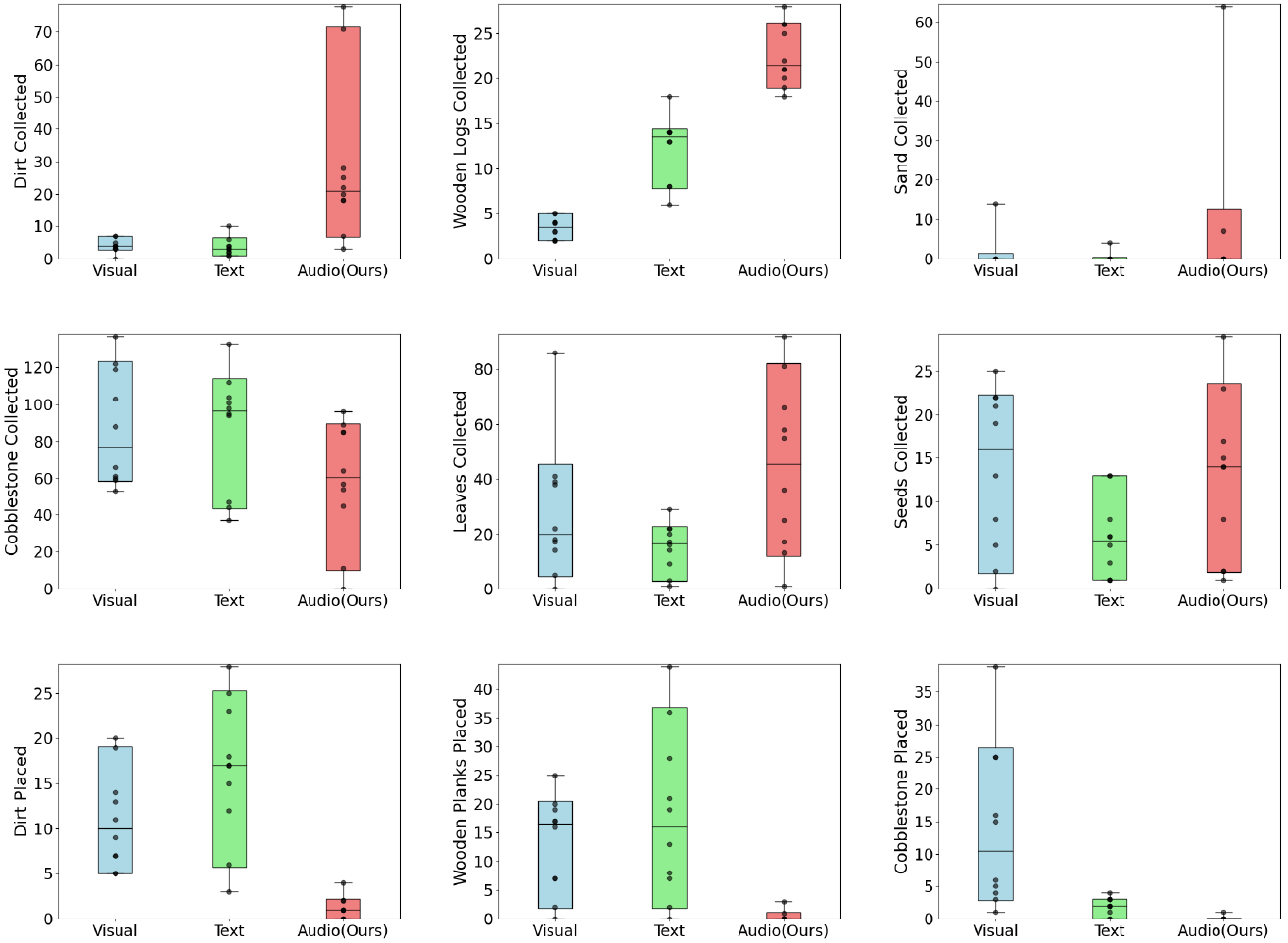}
     \caption{
Performance comparison of audio-conditioned STEVE-1 (created using our proposed methodology) with the original text-conditioned and visual-conditioned STEVE-1 agents. The last row consists of three evaluation with each prompting modality on ``place'' tasks which have ambiguous or underspecified audio prompts (i.e., audio samples for placing dirt and sand sound very similar to the audio samples for digging dirt and sand). Results indicate that audio prompting fails to effectively condition the STEVE-1 policy in these ambiguous or underspecified scenarios. The black bars represent the 10th, 50th, and 90th percentiles, indicating performance spread across the different modalities.
         }
     \label{fig:results}
\end{figure}

\section{Experimental Results}
\label{sec:result}

In our experiments, we aim to answer the following questions:
\begin{itemize}
    \item[1.] How well does our audio-conditioned STEVE-1 agent perform compared to the original text-conditioned and visual-conditioned versions?
    \item[2.] What are the tradeoffs that occur when switching between different prompting modalities?
\end{itemize}

\subsection{Audio-Conditioned STEVE-1}

In Figure \ref{fig:results}, we compare the performance of the audio-conditioned STEVE-1 agent to the original text-conditioned and visual-conditioned STEVE-1 agents \cite{lifshitz2023steve1} on various short-horizon tasks. The audio-conditioned agent performs better than the visual-conditioned agent in four of the six evaluation tasks. Specifically, the audio-conditioned agent collects 6.4$\times$ more wood, 7.25$\times$ more dirt, 5.1$\times$ more sand, and 1.6$\times$ more leaves than the visual-conditioned agent, while it collects 0.9$\times$ the amount of seeds and 0.7$\times$ the amount of cobblestone. When compared to the text-conditioned agent, the audio-conditioned agent collects 1.8$\times$ more wood, 8$\times$ more dirt, 2.2$\times$ more seeds, 17.7$\times$ more sand, and 2.9$\times$ more leaves, while it collects 0.7$\times$ the amount of cobblestone (the only task where the audio-conditioned agent performed worse than the text-conditioned agent). These results indicate that our audio-conditioned agent generally performs better than the original STEVE-1 modalities, which suggests that our proposed methodology is an effective way to extend the prompting modalities of generative agents created using the STEVE-1 approach.

We will make the generated videos available on our website as part of the supplementary material submission.

\subsection{Multi-Modality Tradeoffs}

Our proposed method enables adding new prompting modalities to existing generative agents created with the STEVE-1 approach \cite{lifshitz2023steve1}. In this section, we investigate the tradeoffs that occur when switching between different prompting modalities. In fact, as described below, these positive and negative tradeoffs are a significant motivation for creating such multi-modal agents and extending the prompting modalities of existing agents. That is, creating increasingly multi-modal agents allows them to take advantage of the positive tradeoffs provided by each prompting modality.

\subsubsection{Versatility VS. Performance}
\label{sec:vers-perf}

An important tradeoff to consider between modalities is the ability of each modality to unambiguously express complex instructions. For example, the text modality facilitates communicating complex tasks with decision-making agents (i.e., ``obtain a bed and place it on your roof next to your dog, but don't steal the bed from a village''). On the other hand, such complex instructions would be difficult to communicate with other modalities like audio. Furthermore, beyond specifying complex instructions, it seems that prompting with audio can lead to scenarios where the instruction is too ambiguous or non-specific to be completed. To investigate this, we evaluate the text, visual, and audio versions of STEVE-1 on three different ``placing'' tasks, where the audio is more ambiguous. For example, the audio samples for placing dirt sounds very similar to the ones for digging dirt , and audio samples for placing wooden planks or cobblestone sounds very similar to the audio samples for placing any Minecraft item that is made out of wood or stone, respectively.The last row of Figure 4 shows that audio prompting generally performs much poorer than text and visual prompting, which supports the idea that some tasks are more ambiguous in other modalities (i.e., audio).

However, despite audio being a more ambiguous way to specify some tasks, the audio-conditioned STEVE-1 agent generally performs better than the text and visual counterparts. We hypothesize that this is due to the higher correlation between the audio and video modalities, than exists between text and video. That is, the Text-Video dataset used to train MineCLIP \cite{fan2022minedojo} was sourced from YouTube videos of Minecraft gameplay and their time-aligned captions. Thus, much of the text in the dataset is noisy in the sense that the text is unrelated or uncorrelated to the Minecraft behavior observed in the video (i.e., when someone says ``please like this video''). On the other hand, audio and video are both highly correlated, naturally co-occurring modalities. This means that the audio-video pairs in our Audio-Video dataset are more highly correlated. %

\subsubsection{Prompt Engineering}

In our experiments, prompts for all three modalities were selected with minimal prompt engineering. Thus, since audio-conditioning generally yields better performance, our results suggest that audio prompting might require less prompt engineering than text and visual prompting to perform well. This requires further investigation but could be explained due to the fact that audio samples are usually very similar across different demonstrations of the same task. Thus, the audio embeddings from our Audio-Video CLIP model could be more semantically relevant representations of the task, less affected by small changes in the prompt (i.e., visual embeddings for the same task could be affected by random objects moving in the periphery). This may play a role in the improved audio-conditioning performance. However, more investigation is needed.

\section{Conclusion}
\label{sec:conclusion}

This paper introduces a methodology for extending the prompting modalities of generative instruction-following agents.
By creating increasingly multi-modal agents, singular agents can leverage the advantages and tradeoffs provided by different prompting modalities when completing a task. 
We apply our methodology to the challenging Minecraft domain and extend the existing STEVE-1 agent to follow audio prompts as well. The resulting audio-prompted STEVE-Audio agent outperforms the original text and visual-prompted versions of STEVE-1 on short-horizon item-collection tasks in Minecraft.
Future work should investigate applying this methodology for extending other
generative agents to different sensory modalities and in different domains.

While our method achieves strong results, it has several limitations. First, it requires training a CLIP model \cite{radford2021learning} which takes the new prompting modality as one of its inputs and which requires a dataset of correlated modality-to-modality pairs. 
Second, our approach is limited by the tradeoffs of the prompting modalities to which we extend the agent. For example, it is often harder to specify some tasks using audio prompts compared to text prompts. %

\bibliography{main.bib}  %

\clearpage

\appendix
\section{Dataset and Training}
\label{appendix}

We also augmented our Audio-Video dataset with short 10-20 minute videos that are more specifically relevant to our evaluated set of tasks (an additional two hours in total). That is, we found that using YouTube data exclusively yielded an underperforming audio-conditioned STEVE-1 agent (see Experimental results section below), but including a short amount of more task-specific data significantly improved performance. 

To conform with the specifications of the audio and video encoder models, audio is resampled to a sample rate of $16$ kHz and video is downscaled to a resolution of $160 \times 256$ and set to $32$ frames-per-second, from which we sample 16 evenly distributed frames per sample. We extract audio-video samples from these videos using a sliding window with a length of one second and an overlap of $75\%$. See Table \ref{table:dataset-metrics} and \ref{table:train-hyperparams} for more details.

\begin{table}[!ht]
  \caption{Dataset Metrics}
  \label{table:dataset-metrics}
  \centering
  \begin{tabular}{ll}
    \toprule
    Image resolution            & $160 \times 256$     \\
    Frame rate                  & $32$ fps      \\
    Sample rate                 & $16$ kHz      \\
    Length of samples           & 1 second      \\
    Overlap                     & $75\%$        \\
    Number of train samples     & $303 174$      \\
    Number of test samples      &               \\
    \midrule
    Audio embedding dimension   & $527$         \\
    Video embedding dimension   & $512$         \\
    \bottomrule
  \end{tabular}
\end{table}

\begin{table}[!ht]
  \caption{Training Hyperparameters}
  \label{table:train-hyperparams}
  \centering
  \begin{tabular}{ll}
    \toprule
    batch size          & $1024$        \\
    learning rate       & $0.001$       \\
    Number of epochs    & $100$         \\
    Adam $\beta_1$      & $0.9$         \\
    Adam $\beta_2$      & $0.999$       \\
    weight decay        & $0.01$        \\
    epochs              & $100$         \\
    \bottomrule
  \end{tabular}
\end{table}

\end{document}